\documentclass[conference]{IEEEtran}
\IEEEoverridecommandlockouts
\usepackage{cite}
\usepackage{amsmath,amssymb,amsfonts}
\usepackage{algorithm}
\usepackage[noend]{algpseudocode}
\usepackage{graphicx}
\usepackage[export]{adjustbox}
\usepackage{textcomp}
\usepackage{xcolor}
\usepackage{multirow}
\usepackage{caption}
\usepackage{subcaption}
\usepackage[percent]{overpic}
\usepackage{pgfplots}
\usepackage{url}
\pgfplotsset{compat=1.17}

\def\BibTeX{{\rm B\kern-.05em{\sc i\kern-.025em b}\kern-.08em
    T\kern-.1667em\lower.7ex\hbox{E}\kern-.125emX}}

\begin{document}

\title{
LiDAR Cluster First and Camera Inference Later: A New Perspective Towards Autonomous Driving \\
}

\author{} 
\author{
  \IEEEauthorblockN{
     Jiyang Chen,
     Simon Yu,
     Rohan Tabish,
     Ayoosh Bansal,
     Shengzhong Liu,
     Tarek Abdelzaher, \\ 
     and
     Lui Sha
  }
  \IEEEauthorblockA{
     University of Illinois at Urbana-Champaign,  USA,
     \{jchen185, jundayu2, rtabish, ayooshb2, sl29, zaher, lrs\}@illinois.edu\\
  }
}


\maketitle

\begin{abstract}

Object detection in state-of-the-art Autonomous Vehicles (AV) framework relies heavily on deep neural networks. Typically, these networks perform object detection uniformly on the entire camera/LiDAR frames. However, this uniformity jeopardizes the safety of the AV by giving the same priority to all objects in the scenes regardless of their risk of collision to the AV. In this paper, we present a new end-to-end pipeline for AV that introduces the concept of \textit{LiDAR cluster first and camera inference later} to detect and classify objects. The benefits of our proposed framework are twofold. First, our pipeline prioritizes detecting objects that pose a higher risk of collision to the AV, giving more time for the AV to react to unsafe conditions. Second, it also provides, on average, faster inference speeds compared to popular deep neural network pipelines. We design our framework using the real-world datasets, the Waymo Open Dataset, solving challenges arising from the limitations of LiDAR sensors and object detection algorithms. We show that our novel object detection pipeline prioritizes the detection of higher risk objects while simultaneously achieving comparable accuracy and a 25\% higher average speed compared to camera inference only.

\end{abstract}


\section{Introduction}
\label{sec:intro}

In traditional machine learning pipelines for perception systems, the camera images and the 3D LiDAR point cloud are passed separately through a deep neural network (DNN) for inference~\cite{apollo}.
Following the DNNs, a data fusion component takes the inference results across different DNNs and produces the final object detection results, such as object type, distance, and velocity.
While the input data, such as a camera image, contains objects with varying importance, they are all processed uniformly regardless of their criticality in these pipelines.
Such uniformity introduces significant priority inversion as shown by~\cite{2020removing}. 
In addition, only a portion of the input data may contain helpful information for the AV for collision avoidance~\cite{9460196} or path planning~\cite{Philion2020LearningTE}.
Thus, processing the whole input frame wastes computational resources and delays the processing of higher-risk objects in the scene that may jeopardize the safety of the AV.
Current systems attempt to reduce the severity of this problem by over-provisioning the platforms, which hinders their large-scale deployment to cost and latency-sensitive systems, such as the AVs.
Prior works in the literature partially address this problem by maximizing GPU parallelism with task batching~\cite{2020removing} or resizing~\cite{yigong2021exploring}. However, the prior works make a critical assumption that perfect bounding boxes for the regions of interest in a camera image with corresponding depth information are available.
Our work removes this assumption by using real-world sensor data and addressing the coming unique challenges, such as false-positive and repetitive detection. 


\begin{figure}[t]
    \centering
    \includegraphics[width=\linewidth,keepaspectratio]{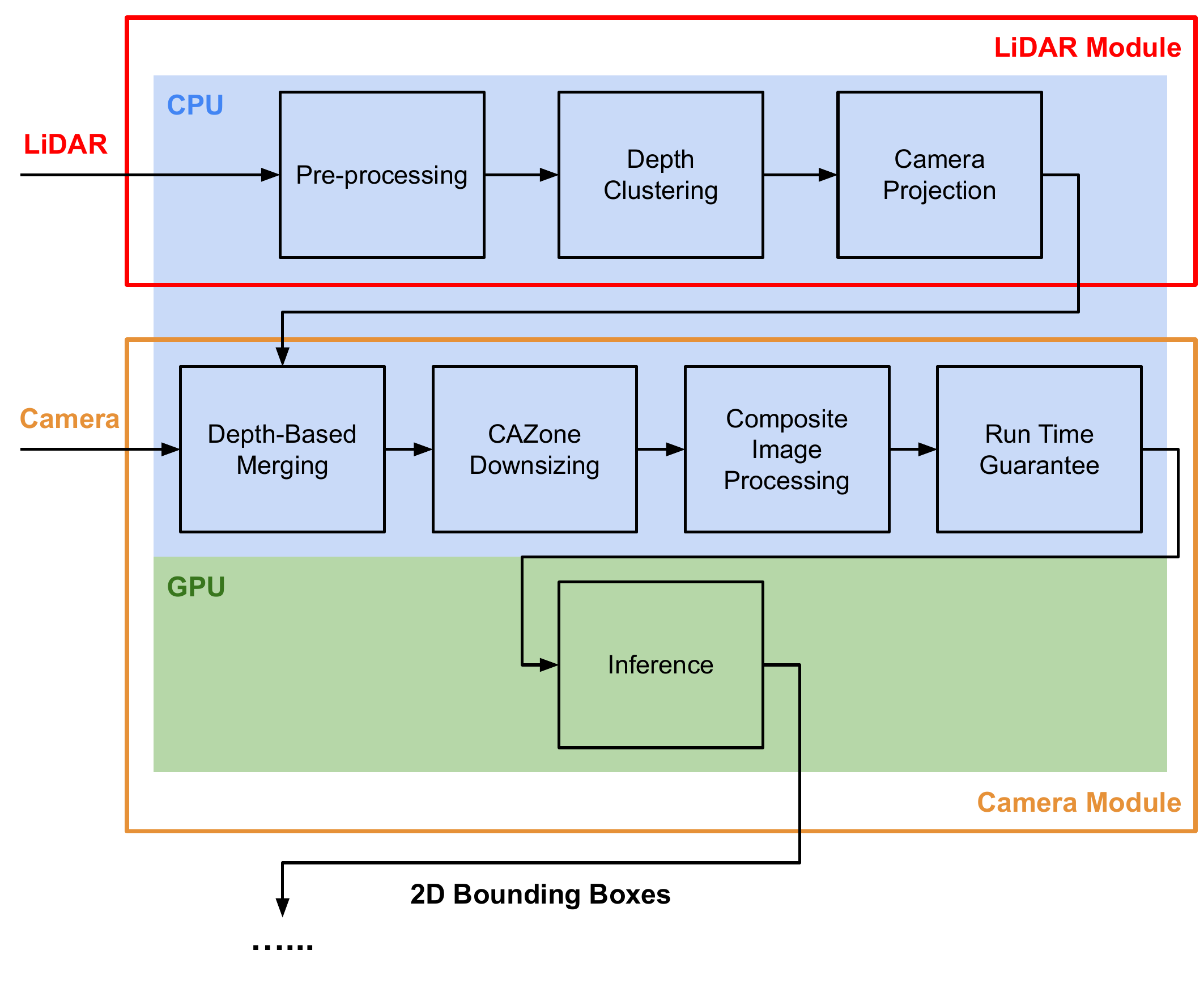}
    \captionsetup{font=small,labelfont=normalsize}
    \caption{LiDAR Cluster First and Camera Inference Later Pipeline.}
    \label{fig:pipeline}
\end{figure}


We propose a novel framework, \textit{LiDAR cluster first and camera inference later}, bringing prioritized object detection to the real world.
An overview is shown in Figure~\ref{fig:pipeline}.
Our proposed pipeline consists of the LiDAR module and Camera module as shown in Figure~\ref{fig:pipeline}.
In the LiDAR module, we employ LiDAR-based Depth Clustering algorithm~\cite{bogoslavskyi17pfg} 
for performing fast segmentation on 3D point clouds, which produces \textit{3D clusters} for detected objects.
The 3D clusters are then projected to the camera image as rectangular boxes, which we call \textit{2D clusters}.
Each 2D cluster, ideally, encloses one physical object.
2D clusters also have an associated depth which indicates how far the object is from the AV.

The output from the LiDAR module is then passed to the Camera module.
A direct consequence of using real-world data is that the output from the LiDAR module contains false positive 2D clusters.
The 2D clusters can also be over-segmented, i.e., scattered across the scene or densely overlapping.
These challenges are addressed within the Camera Module by a depth-aware merging algorithm ($\S$\ref{sec:merge_component}) that merges 2D clusters into Collision Avoidance Zones, or \textit{CAZone}.
As the name describes, \textbf{CAZone} is defined as a region where there are potential obstacles the vehicle should try to avoid.
We further propose a composite image approach to process these CAZones efficiently, improving the end-to-end response time of the system.

Therefore, the main contributions of this work are:

\begin{itemize}
    \item We use real-world LiDAR data and an existing fast Depth Clustering algorithm~\cite{bogoslavskyi17pfg} to assign criticality information to different regions in a camera image frame.
    \item We introduce a depth-based merging algorithm to combine LiDAR 2D cluster into CAZones for the more accurate enclosing of potential objects in a camera frame.
    \item We propose a depth-aware First-Fit Decreasing Height (FFDH) algorithm to group CAZones into composite images and use depth-based optimizations and GPU profiling to bound priority inversion and run time. 
    \item Finally, we provide an end-to-end evaluation of the system using Waymo Open Dataset and show that our approach has better run time while maintaining the same level of accuracy as compared with processing the full camera frame.
\end{itemize}

\section{The LiDAR Module}
\label{sec:lidar_module}

The LiDAR module takes the 3D point cloud from the LiDAR sensor, performs efficient point cloud clustering and segmentation, and projects the resulting 3D clusters onto the 2D camera images. The module has three components, each performing distinct tasks, as shown in Fig.~\ref{fig:pipeline}.

\textit{Pre-processing Component.}
The Pre-processing component takes the raw 3D LiDAR point cloud and limits the horizontal field of view of the point cloud ($360^{\circ}$) to the field of view of the camera ($50.4^{\circ}$), as provided by the Waymo Open Dataset~\cite{sun2020scalability}.

\textit{Depth Clustering Component.}
The Depth Clustering component segments the point cloud into various clusters representing the individual objects in the scene. This module is based on the Depth Clustering algorithm and implementation provided by Bogoslavskyi and Stachniss~\cite{bogoslavskyi17pfg}.
Depth Clustering works based on thresholding the level of depth discontinuities among the points in the point cloud.
For example, suppose two adjacent points in the point cloud have significant depth disparity. In that case, the two points likely belong to a different object and vice versa if the depth disparity is low.


\textit{Camera Projection Component.}
The Camera Projection component takes the 3D clusters from the Depth Clustering component and projects them onto the camera images as the 2D clusters. The component extracts essential camera calibration information from the Waymo Open Dataset, such as the extrinsic and intrinsic matrices, and performs a standard camera projection on all the points in the 3D clusters. Each 2D cluster is associated with a depth, i.e., the distance between the closest point in the cluster and the AV, and it should enclose a single object in the scene. This depth information is used later in the Camera module to determine the criticality of the objects in the scene. The generated list of 2D clusters with depth is then sent to the Camera module.

\textit{Limitations.}
The accuracy of the LiDAR module decreases with noise in sensor data, e.g., due to fog, rain, snow, etc.
Therefore, we limit the evaluation in this work to highway scenes with sunny and clear weather conditions and assume that all relevant objects are enclosed by the 2D clusters generated by the LiDAR module.
Adverse weather conditions are outside the scope of this paper and will be addressed in future work.

\section{The Camera Module}
\label{sec:camera_module}

%
The Camera module addresses the practical challenges of false positives and over-segmentation within the 2D Clusters.
We devise a new approach that utilizes the 2D Clusters to improve DNN inference time without compromising detection accuracy.
The module also distinguishes critical and non-critical objects in the corresponding camera image.
Objects closer to the AV are more critical, and the corresponding CAZones are guaranteed to be completely processed by the end-to-end pipeline within a preset run time threshold.

\subsection{2D Cluster Merging}
\label{sec:merge_component}

\begin{figure}[t]
    \centering
    \includegraphics[width=.9\linewidth,keepaspectratio]{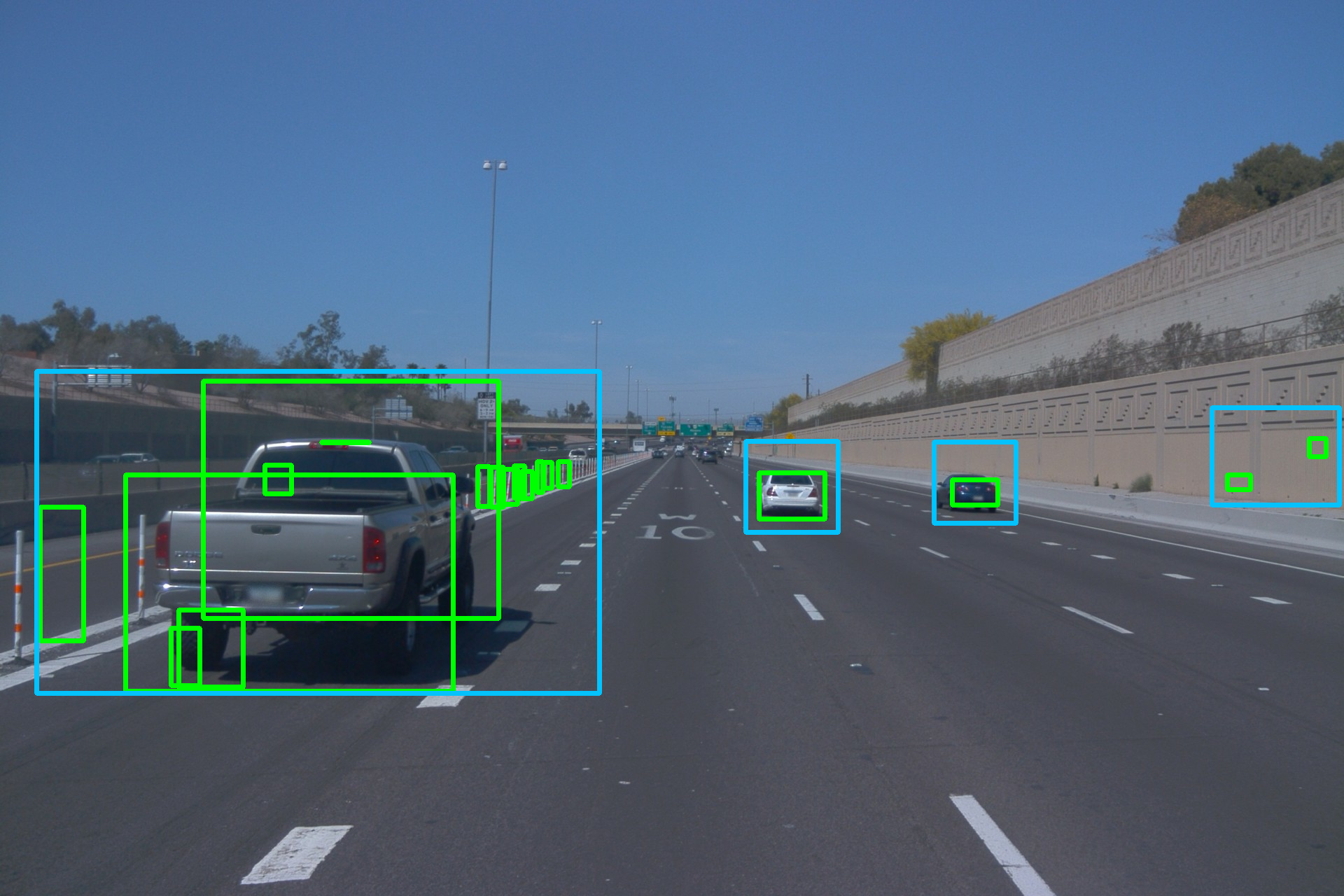}
    \caption{2D Clusters \textit{vs.} CAZones. The green boxes are 2D clusters and the blue boxes are CAZones.}
    \label{fig:cluster}
\end{figure}

We discovered many practical challenges with the 2D clusters produced from the LiDAR module by employing a real-world dataset with the Depth Clustering algorithm. 
The number of generated 2D clusters is more than the actual objects in the scene due to the following reasons:
\begin{enumerate}
    \item There can be false-positive 2D clusters that do not enclose any objects.
    \item Since it does not perform classification, the Depth Clustering algorithm can thus over-segment the clusters, as shown in Figure~\ref{fig:cluster}.
    \item 2D clusters tend to be smaller than the object's size in the scene due to the limitations of LiDAR sensors.
    \item Sometimes, the 2D clusters only enclose the objects partially, which could result in inaccurately created CAZones and thus affect the inference performance of the Camera module later down in the pipeline.
\end{enumerate}
As a result, inferencing the 2D clusters with a DNN would result in inefficiencies and poor accuracy.


The purpose of the 2D Cluster Merging component is to take the raw 2D clusters to form a list of CAZones such that all physical objects in the frame are enclosed.
As a first step, we use inverse depth-based inflation to increase the size of all 2D clusters.
2D clusters closer to the AV are inflated less, while the farther-away 2D clusters are inflated more.
This inverse relationship is derived from LiDAR beam dispersion and divergence increases with distance, increasing the error in 2D clusters for farther objects.

We develop a depth-based merging algorithm, Algorithm~\ref{alg:merge}, to create CAZones out of 2D clusters.
A pair of 2D clusters is merged under the following two conditions.
First, the 2D clusters are merged if they are close, i.e., similar depth and have a small pixel distance in the camera image, as in line 6 in Algorithm~\ref{alg:merge}.
Meeting the above condition implies that the 2D clusters represent a zone that the AV cannot physically pass through, i.e., a CAZone.
Before checking whether the two clusters are close in camera image pixel distance, the width and height of both clusters are increased by a margin.
The margin is linear to their depth, as in line 15 in Algorithm~\ref{alg:check_close}, and the parameter $a$ is determined empirically.
Two clusters, with their margins, are determined to be close if their intersection over union (IoU) is greater than 0.1.
Second, two clusters are merged regardless of their depths if they have an IoU greater than $0.3$.
Such 2D clusters likely represent occluded objects.

The depth-based merging algorithm finishes when no 2D cluster pairs can be merged further.
When a pair of 2D clusters is merged, the new cluster inherits the minimum depth of the original cluster pair.
The final product of the Algorithm~\ref{alg:merge} is a list of non-overlapping CAZones, each of which represents potential physical objects in the scene. 
These CAZones are assigned high or low priority based on a safety distance threshold.
The safety distance shall be derived from the relevant driving standards and safety laws.
In our case, as an example, we choose the safety distance based on UK's highway code~\cite{ukcode} which prescribes safe inter-vehicle distance as a function of the vehicle speed.
All CAZones within this safety distance are assigned high priority (HP) while others are assigned low priority (LP).


\begin{algorithm}[t]
    \caption{Depth-Based Merging}
    \begin{algorithmic}[1]
    \State Input: cluster set $\mathcal{C}$
    \State Merged = True 
    \While {Merged}
        \State Merged = False
        \For{$C_1$ in $\mathcal{C}$}
            \If {$\exists C_2$ in $\mathcal{C}$ s.t. $check\_close(C_1, C_2)$ is True}
                \State Merge $C_2$ into $C_1$;
                \State Merged = True
            \ElsIf {$\exists C_2$ in $\mathcal{C}$ s.t. $IoU(C_1, C_2) > 0.3$ }
                \State Merge $C_2$ into $C_1$;
                \State Merged = True
            \EndIf
        \EndFor
    \EndWhile 
    \State Return $\mathcal{C}$
    \vskip 2mm

    \Function{check\_close}{cluster $C_1$, cluster $C_2$}
    \If {$C_1\_depth$ and $C_2\_depth$ are within length L}
        \State Increase $C_1$ size by  $a \cdot C_1\_depth$
        \State Increase $C_2$ size by  $a \cdot C_2\_depth$
        \If { $IoU(C_1, C_2) > 0.1$}
            \State Return True
        \EndIf
    \EndIf
    \State Return False
    \EndFunction     
    \end{algorithmic}
\label{alg:merge}    
\label{alg:check_close}
\end{algorithm}

\subsection{CAZone Downsizing}

For this component, the input lists CAZones with depth that encloses potential physical objects in the scene. 
These CAZones retain the complete pixel information from the original image and can be larger than the input size required by the deep neural network to detect objects accurately.
Thus, in existing AV stacks, images are first downsized uniformly and sent to the perception module. For example, YOLOv3-608~\cite{redmon2018yolov3} downsizes various input sizes to 608 x 608 images.
This downsizing applies to all objects by the same factor, regardless of their properties and criticality.
As a result, the closer objects might be unnecessarily large while, simultaneously, objects that are farther away become too small and thus difficult to identify.
The closer objects can be downsized more without affecting their detection quality, while objects that are farther should be downsized less, so it becomes easier for the perception module to detect them.
Therefore, we propose variable downsizing for objects based on their distance to the vehicle, improving the detection quality for farther objects with minimal impact on the detection quality for closer objects.
In this way, we can adjust the pixel information for objects of varying depth using a linear relation to determine the downsize factor as shown in Equation~\ref{eq:downsize}:
\begin{equation}
    D = D_0 - b \cdot depth
\label{eq:downsize}
\end{equation}
where $D_0$ is a base downsizing factor, and b is a constant bias for adjusting the effect of depth.
For our pipeline, we choose $D_0$ to be 3 $(\approx \frac{1920}{608})$ and $b$ to be $\frac{2}{75}$.
The above choice means the closest object (0 m depth) will have about the same level of downsizing as in YOLOv3-608 (D = 3) while the farthest away objects (75 m depth) will not be downsized (D = 1).

\subsection{Fitting CAZone to Composite Image}

\begin{figure}[t]
\captionsetup[subfigure]{justification=centering}
\centering
        \centering
        \includegraphics[width=\linewidth,keepaspectratio]{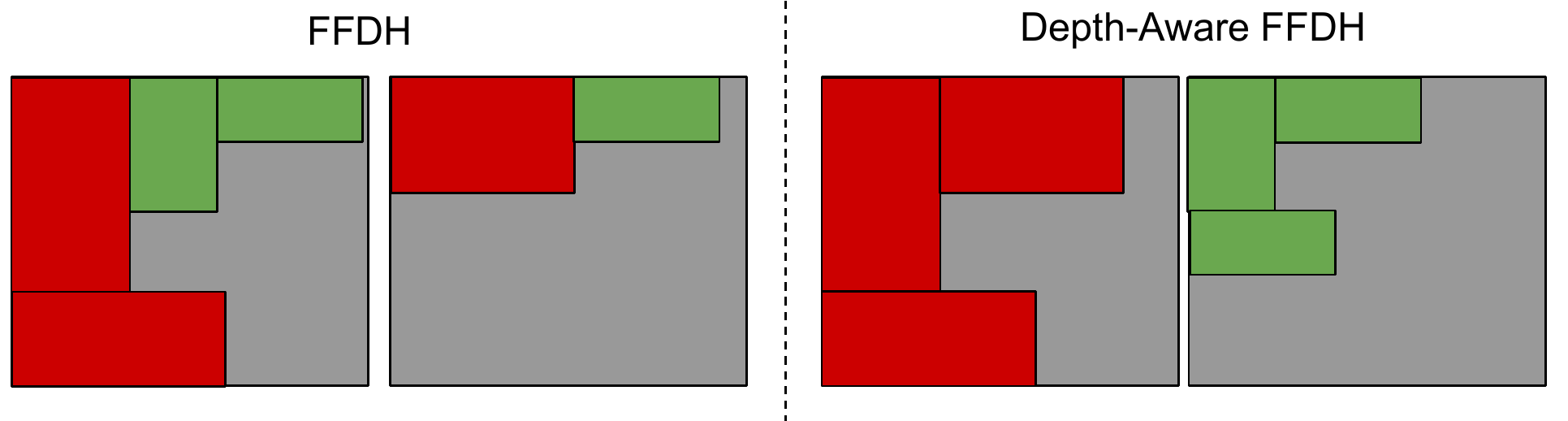}
        \caption{FFDH \textit{vs.} Depth-Aware FFDH}
    \hfill
    \label{fig:ffdh}
\end{figure}

The input to this component is a list of variably downsized CAZones.
A limitation of the platform becomes relevant here.
Most widely available embedded GPUs cannot concurrently execute different compute kernels.
Batching of the same input size, executing multiple inputs with the same compute kernel, is possible.
Therefore, to better utilize the GPU, rather than simply passing the CAZones to the inference component, we propose a composite image approach such that different CAZones can be grouped and executed together.
The composite image approach is designed to be flexible, adapting to the parallelism capabilities of any given GPU.
As shown in Figure~\ref{fig:ffdh}, a composite image is created by fitting CAZones onto an empty square canvas.
We choose only one canvas size $S$ for a frame for maximum parallel execution throughput on the embedded GPU.
Depending on the number of CAZones in the frame, there might be multiple canvases with the chosen size $S$.
We further define that composite image is HP if it has at least one HP CAZone, and LP if otherwise.


\begin{figure}[t]
\centering
\includegraphics[width=.5\linewidth,keepaspectratio,frame]{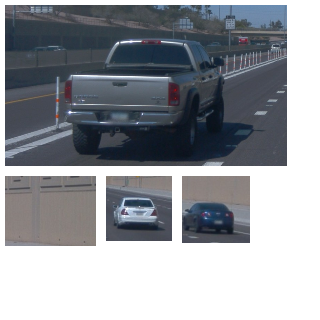}
\caption{A example of a composite image formed by CAZones.
}
\label{fig:composite}
\end{figure}

For the composite images to run efficiently, it is essential to 1) choose an optimal canvas size $S$; 2) find an optimal fitting for all CAZones such that the number of composite images is minimal.
We determine the canvas size based on the $max(width,\ height)$ of all CAZones and then round it up to the nearest multiple of 32, which is compatible with the input size requirement of YOLOv3, thus making sure even the largest CAZone can fit inside the canvas.
For fitting all CAZones, we propose a depth-aware First-Fit Decreasing Height (FFDH) algorithm, which is a modified version of the original FFDH algorithm~\cite{coffman1980performance}, a classical approximation solution for the two-dimension bin packing problem.
The original algorithm assumes all items are sorted by non-increasing height.
The next item is packed, left-justified, on the first level (initially, the top of the bin) in which it fits.
If no such level is found, a new level is created, and the item fits there.
The depth-aware FFDH will first sort HP and LP CAZones by height separately to form two sorted lists to reduce priority inversion.
Next, the depth-aware FFDH will first fit all CAZones from the HP list, produce HP composite images, and then the LP CAZones.
If there is space available in HP composite images, the LP CAZones can be inserted into HP composite images if they fit.
If not, new LP composite images are created.
In the end, the depth-aware FFDH algorithm produces a list of HP and LP composite images containing all the CAZones in the frame.
To improve detection accuracy, the composites images are adjusted to separate the fitted CAZones with gaps among each other and away from the canvas edges.
The background color for the canvas is chosen to be treated as null pixels by YOLOv3.

\subsection{Run Time Guarantee for HP Composite Images}

The purpose of this module is to reduce priority inversion and provide a guarantee that the run time of the HP composite images does not exceed a preset threshold.
We profile the run time of the YOLOv3 under different input sizes and batch sizes, and part of the results is shown in Table~\ref{tab:gpu_profile}.
The composite images from each frame are checked against the table to see if overrun will occur if all composite images are run together.
HP composite images are always run while guaranteeing the preset threshold.
The LP composite images are scheduled in a better-effort manner, as typical for soft real-time tasks.
If an overrun would occur upon running the complete input list, LP composite images are gradually removed from the input list until the run time requirement can be met. 
Suppose the input composite images still exceed the run-time requirement according to Table~\ref{tab:gpu_profile} after all LP composite images are removed. In that case, the size of the remaining HP composite images is reduced to the maximum size that would still meet the threshold given the count of HP composite images in the input list, i.e., the batch size in Table~\ref{tab:gpu_profile}.
For example, there are 3 HP and 2 LP composite images with size 288x288 for a frame, and the run-time target is set to 140 ms.
According to Table~\ref{tab:gpu_profile}, the composite images will overrun, so the 2 LP composite images are removed first.
However, three images with 288x288 are still exceeding 140 ms 
so all three images are downsized to 256x256.
In an extreme case of heavy traffic, there could be a case that CAZones cover most, if not all, of the camera image.
In such scenarios, we process the full-frame without further pre-processing, at a size according to Table~\ref{tab:gpu_profile}, instead of using the CAZones.
This still guarantees the run time threshold, sacrificing a bit of the detection accuracy.

\begin{table}[t]
\centering
\caption{GPU Run Time (ms) \textit{vs.} Batch Size \& Input Size.}
\label{tab:gpu_profile}
\begin{tabular}{|c|c|c|c|c|c|c|c|}
\hline
\begin{tabular}[c]{@{}c@{}}input\\ batch\end{tabular} & 192   & 256   & 288   & 352   & 416   & 512   & 608   \\ \hline
1                                                     & 75 & 76 & 90 & 99 & 115 & 127 & 173 \\ \hline
2                                                     & 84 & 95 & 124 & 140 & 169 &       &       \\ \hline
3                                                     & 96 & 115 & 148 & 180 &       &       &       \\ \hline
4                                                     & 111 & 133 & 186 &       &       &       &       \\ \hline
5                                                     & 124 & 153 &       &       &       &       &       \\ \hline
6                                                     & 135 &       &       &       &       &       &       \\ \hline
7                                                     & 145 &       &       &       &       &       &       \\ \hline
\end{tabular}
\end{table}

\subsection{Inference}
For the Camera module's Inference component, we use YOLOv3~\cite{redmon2018yolov3}, a popular deep neural network for camera-based object detection. 
Any neural network that accepts variable input sizes can be used in our framework, such as all Fully Convolutional Networks (FCN) like YOLOv3 and ResNet~\cite{he2016deep}. The use of ResNet has been demonstrated in~\cite{2020removing}.

\section{Evaluation}
\label{sec:eval}


We use YOLOv3 DNN as the baseline for comparison to evaluate our system, which takes complete camera images as input and outputs 2D bounding boxes for the detected objects. In the experiments below, we show that our approach performs detections faster with a comparable level of accuracy of full-frame.
We perform our evaluations on the COTS NVIDIA Jetson AGX Xavier \footnote{https://developer.nvidia.com/embedded/jetson-agx-xavier-developer-kit} platform designed for intelligent edge devices. The platform contains 8 ARM 64-bit CPUs, a 512-core Volta GPU, and has 32 GB of main memory. During our experiments, we set the power mode of the platform to MAXN to minimize variability due to power management features.

We performed the evaluation on the Waymo Open Dataset \cite{sun2020scalability} which contains various sensor data collected from the Waymo's self-driving fleet in diverse driving conditions. In the dataset, each of the driving segments is 20 seconds long, with LiDAR and camera data collected at 10 Hz. The data between LiDAR and the camera is well-synchronized and is labeled "vehicle," "pedestrian," "cyclist," and "sign" for both sensors. In this work, we use data from the front camera and the top LiDAR, which has a maximum effective range of 75 m.
To meet our assumption on the limitation of the LiDAR sensor as stated in Section~\ref{sec:lidar_module}, we only use segments on the highway, clear sunny days with an average speed larger than 29 m/s. We have found ten segments that meet the criterion, and we used 6 out of 10 for testing and parameter-tuning. The remaining four are used for the evaluation.



YOLOv3-608 takes 200 ms to produce bounding boxes given a full camera image input, including any pre-processing and post-processing time.
We set a goal of having a 20\% lower end-to-end time than YOLOv3-608, i.e., 160 ms. Based on empirical evaluation of LiDAR and Camera modules (Figure~\ref{fig:runtime}), we determined a worst-case budget of 140 ms for the Camera module inference component, i.e., YOLOv3 with variable input size.
Various input sizes between 128x128 to 512x512 and batch combination sizes that meet the criterion of run time less than or equal to 140 ms are acceptable, as shown in Table~\ref{tab:gpu_profile}. 
For the full-frame approach, we use two setups. 
The original image with resolution 1920x1280 is downsampled to an input size of 608x608 (downsampling ratio 3.15), representing the default setup.
We use 608 since it is the largest size evaluated by YOLOv3~\cite{redmon2018yolov3}, yielding the highest mAP.
We also chose an input size of 512x512 because it has a comparable run time with the composite image approach, and we want to see if our pre-processing optimizations have benefits over simply downsizing the full frame to a smaller size.

Due to several reasons, the original camera ground truths from the Waymo Open Dataset cannot be used directly, and ground truths for objects beyond the LiDAR range (75 m) are ignored.
First, we argue that these objects are not required to be detected consistently as the large distance makes them have a negligible collision risk to the AV.
Second, there exists a limitation of the Waymo Open Dataset where camera labels are not paired with Lidar labels.
Thus, we first match camera and LiDAR labels, using the largest IoU, with a 50\% threshold, as the selection criteria. Only labels where such matches could be found are included, as also done in prior works based on the same dataset~\cite{9460196}.
Finally, given our focus on highway scenes, only ground truths for vehicles are included. 
All types of vehicles detection results given by YOLOv3, including cars, motorbikes, trucks, are classified as vehicles in the detection outputs.

\subsection{Detection Run Time Comparison}

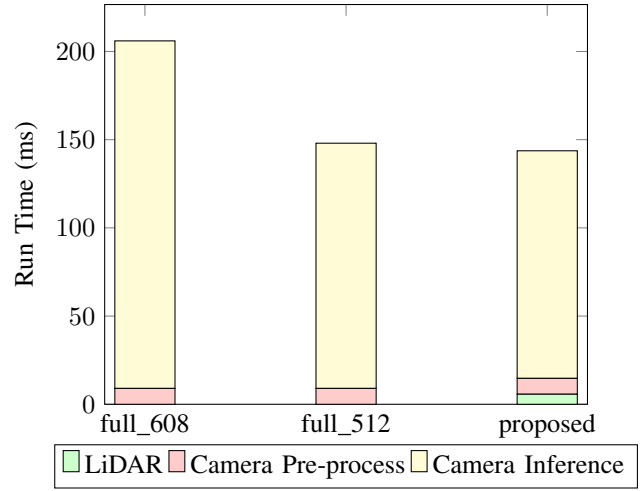
\begin{figure}[t]
\centering
\begin{tikzpicture}[scale=1]
    \pgfplotsset{
        selective show sum on top/.style={
            /pgfplots/scatter/@post marker code/.append code={%
                \ifnum\coordindex=#1
                   \node[
                   at={(normalized axis cs:%
                       \pgfkeysvalueof{/data point/x},%
                       \pgfkeysvalueof{/data point/y})%
                   },
                   anchor=south,
                   ]
                   {\pgfmathprintnumber{\pgfkeysvalueof{/data point/y}}};
                \fi
            },
        },selective show sum on top/.default=0
    }

    \begin{axis}[
    width=8cm,
    ybar stacked, ymin=0,  
    ylabel={Run Time (ms)},
    bar width=8mm,
    symbolic x coords={full\_608, full\_512, proposed},
    xtick=data,
    legend style={at={(0.5,-0.10)},
      anchor=north,legend columns=-1},
    ]
    \addplot [fill=green!20, selective show sum on top] coordinates {
        ({full\_608},0)
        ({full\_512},0)
        ({proposed},5.7)};
    \addplot [fill=red!20] coordinates {
        ({full\_608},9)
        ({full\_512},9)
        ({proposed},9)};
    \addplot [fill=yellow!20,selective show sum on top/.list={0,1,2}] coordinates {
        ({full\_608},197)
        ({full\_512},139)
        ({proposed},129)};
    \legend{LiDAR,Camera Pre-process, Camera Inference}
    \end{axis}
\end{tikzpicture}
\caption{Run Time Comparison.}
\label{fig:runtime}
\end{figure}

We evaluate the run time of the proposed pipeline against YOLOv3 for the validation segments.
As would be the case for a real platform, we assume the LiDAR point cloud and the camera image data are already loaded to the main memory and available for execution. As a result, the time of loading the dataset from the file to the main memory is not included.
The average run time data collected from the validation datasets is shown in Figure~\ref{fig:runtime}.
The Lidar module takes 5.7 ms to run on average.
Components in the Camera module, excluding the inference component, takes 9 ms in total for all three approaches, which means the overhead of our proposed optimizations is negligible.
The run time of the inference component includes running YOLOv3 on GPU and the post-processing of the bounding boxes.
The inference component run-time for the three approaches is 197 ms, 139 ms, and 129 ms, respectively. 
Even with the Lidar module processing the additional LiDAR data, our proposed pipeline is still faster than both full-frame setups.

\subsection{Detection Accuracy Comparison}

The average precision (AP) for the three approaches on all validation datasets is shown in Table~\ref{tab:ap_compare}.
No ground truths matching the definition of $AP_H$ were found. So we only include $AP_M$ and $AP_L$ in Table~\ref{tab:ap_compare}, as defined by COCO Dataset\footnote{https://cocodataset.org/\#detection-eval}.
Since we have retained more pixel data for farther away objects, an effect of the variable downsampling, our proposed approach improves upon $AP_M$ of YOLOv3-608.
This also contributes to the higher score for $AP_{50}$.
Thus our method has comparable, in some cases better, detection accuracy to both full-frame setups.



\begin{table}[t]
\centering
\caption{Average Precision Comparison.}
\label{tab:ap_compare}
\begin{tabular}{|c|c|c|c|c|}
\hline
          & $AP_{50}$ & $AP_{75}$ & $AP_{M}$  & $AP_{L}$   \\ \hline
full\_608 & 95.32\% & 90.33\% & 90.75\% & 99.64\% \\ \hline
full\_512 & 94.51\% & 88.13\% & 89.21\% & 99.52\% \\ \hline
proposed & 96.25\% & 88.59\% & 92.73\% & 99.52\% \\ \hline
\end{tabular}
\end{table}

\section{Related work}
\label{sec:related}

Researchers in the community have developed both Deep Learning (DL) and classical (non-DL) methods to address LiDAR 3D point cloud segmentation and feature extraction problem. DL approaches \cite{lei2020seggcn, engelcke2017vote3deep} require designing and creating complex network architectures to learn sophisticated 3D features for effective segmentation and classifications. This results in inferior run time performance (hundreds of milliseconds) and is thus not suitable for real-time applications that deploy low-end embedded computers. 
Moreover, classifying the 3D point cloud, which is inherently performed by DL architectures, is not required for the proposed framework. 
Similarly, non-DL approaches \cite{douillard2011segmentation, himmelsbach2010fast} often employ grid-based segmentation, which involves building 3D cubic grids and computing 3D features inside the grids. Such methods are considerably time-consuming (over 100 ms) and suffer from inaccuracy in the segmentation along the grid direction.

Liu et al.~\cite{2020removing} described the existence of priority inversion in object detection pipelines and proposed solutions.
It was further extended by using image resizing to find the optimal input size for GPU batching~\cite{yigong2021exploring}.
However, they both assumed ideal LiDAR cluster boxes being available.
Our work is thus unique in creating an end-to-end pipeline with real LiDAR data.
Merging per-camera frames into composite images for improving GPU throughput was proposed in~\cite{yang2019re}.
They considered only merging full frames in a group of four which lacks the flexibility of our approach.
To improve the real-time aspect of DNN, resource sharing and layer-aware DNN approximation to guarantee the run time of DNN workloads for deterministic run time for AI pipelines has been proposed in~\cite{bateni2018apnet}.
Lee et al.~\cite{lee2020subflow} proposed a strategy to dynamically change the DNN subnet to meet timing constraints while ensuring performance.
Our approach, on the other hand, does not require changes to the DNN structures.

\section{Conclusion and Future Work}
\label{sec:conclusion}

In this work, we presented a novel end-to-end framework that combines LiDAR-based clustering with camera-based inference for reducing priority inversion in the current perception system for autonomous driving.
Our presented approach exhibits a faster run time compared to the popular object detection pipeline while providing a comparable detection accuracy.
As a part of our evaluation, we have considered limited scenarios such as perfect weather conditions and highways only. Moreover, the evaluation has been done on a limited dataset, proving its efficacy albeit in limited driving conditions.
Therefore, as our future work, we will extend the capability of the proposed system to handle heavy traffic, city driving, and adverse weather conditions.
%
%
%
Although we focus on the AV application in this work, our solution is generally applicable to any object detection pipeline where input regions are not uniformly important.

\bibliographystyle{IEEEtran}
\bibliography{ref}

\end{document}